# Experimental investigation of trans-scale displacement responses of wrinkle defects in fiber reinforced composite laminates

Li Ma[1]*, Shoulong Wang[1], Changchen Liu[1], Ange Wen[2], Kaidi Ying[1], Jing Guo[1]
1. Institute of Applied Mechanics, Zhejiang University of Technology, 310023
2. Institute of Process Equipment, Zhejiang University, 310027

**Abstract** Wrinkle defects were found widely exist in the field of industrial products, i.e. wind turbine blades and filament-wound composite pressure vessels. The magnitude of wrinkle wavelength varies from several millimeters to over one hundred millimeters. Locating the wrinkle defects and measuring their responses are very important to the assessment of the structures that containing wrinkle defects. A meso-mechanical modeling is presented based on the homogenization method to obtain the effective stiffness of a graded wrinkle. The finite element simulation predicts the trans-scale response of out-of-plane displacement of wrinkled laminates, where the maximum displacement ranges from nanoscale to millimeter scale. Such trans-scale effect requires different measurement approaches to observe the displacement responses. Here we employed Shearography (Speckle Pattern Shearing Interferometry) and fringe projection profilometry (FPP) method respectively according to the different magnitude of displacement. In FPP method, a displacement extraction algorithm was presented to obtain the out-of-plane displacement. The measurement sensitivity and accuracy of Shearography and FPP are compared, which provides a quantitative reference for industrial non-destructive test.

# 1 Introduction

Carbon fiber reinforced polymers (CFRP) have been widely used in the field of aerospace, transportation and energic industry for the significant advantages of high specific modulus and strength, corrosion and fatigue resistance [1], etc. However, defects such as pores, inclusions, wrinkles and delaminations are inevitably introduced into the CFRP structures during the heating and curing process, which seriously weakens the strength, stiffness and fatigue life of composite components [2-5]. The fiber reinforcement in composites may have breakage, wrinkling and waviness during manufacture, where the wrinkle defects, as well as waviness can exist in macro- and meso-scales. Fig.1 indicates Shearography (Speckle Pattern Shearing Interferometry) technique may be an approach for detecting defect responses in multiple scales.

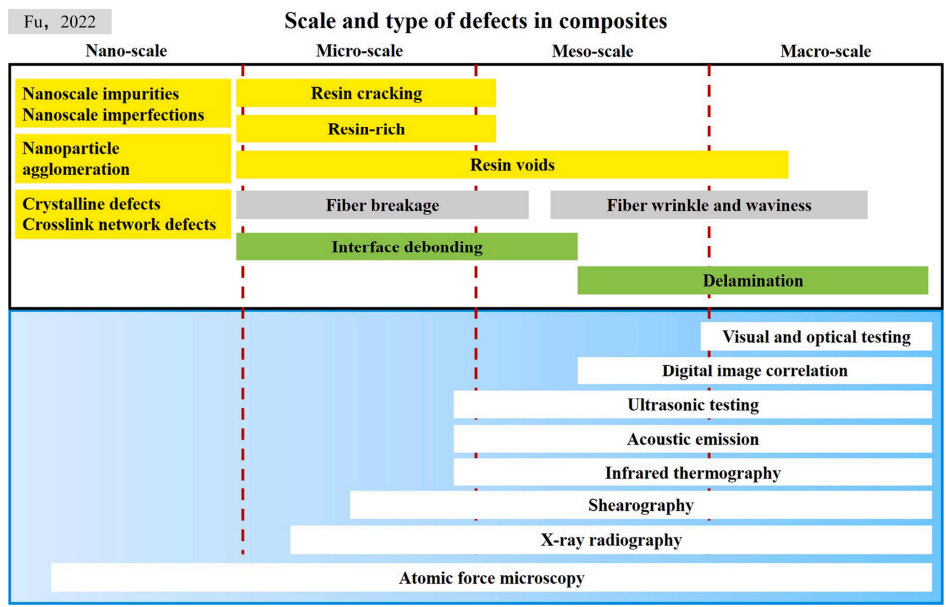

Fig. 1 Scale and type of defects in composites and the available detection methods[6]

Wrinkle defects were frequently found in wind turbine blades, Chen [7] reported a wrinkle with length of 40mm and height of 5mm that caused the damage of blade. The fatigue and post-fatigue crack growth of a wrinkled thick composite laminate that was cut out from a wind turbine blade was investigated, where CT (Computerized Tomography) scanning provided the detail distribution of fiber misalignment angles and the damage was characterized by a Mode-II dominated fracture [9-10].

It was found the measured compressive strength approximately decreases by 5% with a degree increment of fiber misalignment angle [11] and the tensile strength has a 16% reduction with multiple wrinkle defects [12]. Comparing the two severity types of wrinkles, the fatigue lifetime for equal external loading differs by approximately two decades. This difference was also captured by the numerical predictions [13].

The stiffness degradation and the cracking during fatigue test of composite laminates that were embedded with artificial wrinkle defects were further investigated [14,15], where Digital Image Correlation (DIC) and the other optical image analysis method provides the experimental insights into the failure mechanism.

**Table 1 Dimensional sizes of wrinkles**

| Thickness /mm | Amplitude /mm | Wave length $\lambda$/mm | Structural type | Author |
|---|---|---|---|---|
| 15 | 5 | 60 | Wind turbine blade | Miao[8] |
| 43 | 5 | 40 | Wind turbine blade | Chen[7] |
| 0.5 | 1.1 | 15.9 | Laminates | Mizukami[18] |
| 2 | 0.1~0.15 | 2 | Laminates | Wilhelmsson[11] |
| 3 | 0.5±0.04 | 9.6±1 | Laminates | Calvo[12] |
| 3 | 1 | 5 | Laminates | Bloom[16] |
| 4 | 0.18~0.79 | 3 | Laminates | Hallander[17] |
| 6.06 | 1 | 8 | L-shape specimen | Naderi[19] |
| 9 | 2 | 15 | Laminates | Mendonça[13] |
| 14~15.5 | 1~3.2 | 4~40 | Laminates | Davidson[15] |
| 30 | 8~11 | 20~40 | Laminates | Spencer[14] |

Table 1 collects the dimensional sizes of the wrinkles in references [7-8,11-19]. For the wrinkles exist in large structure such as wind turbine blade and filament-wound composite pressure vessels [20], the maximum wave length reaches up to 126.8mm, see Fig.2. For the wrinkles in laminates the wave length ranges from 2mm to 40mm with height from 0.1mm to 11mm, and the ratio of amplitude to wavelength is usually 0.05~0.55.

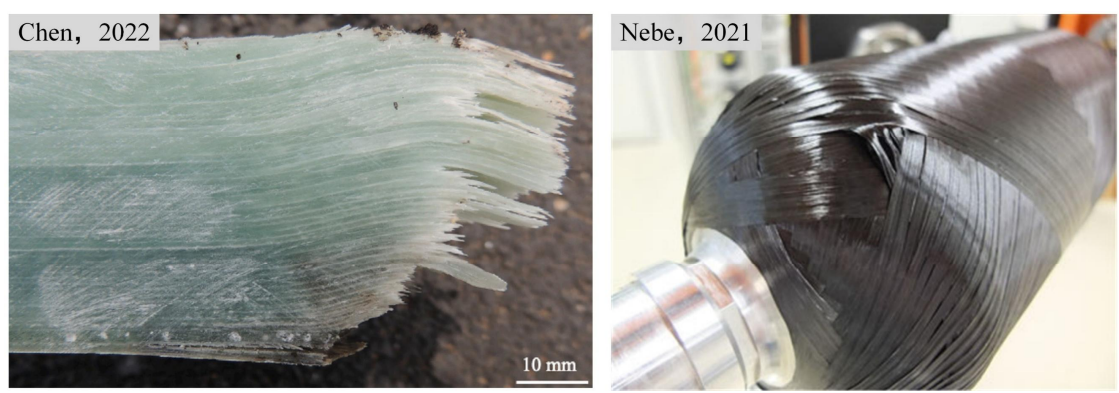

Fig. 2 Wrinkle in wind turbine blade and filament-wound composite pressure vessel

Shearography is a full-field speckle interferometric technique that has a high sensitivity to displacement derivatives and has been used in the non-destructive testing for composites [23]. It was used for real-time in-situ imaging of damage evolution in carbon fiber composites and obtained the minimum 1.3nm of out-of-plane displacement, which shows better defect imaging ability than DIC [24]. The size of the smallest detectable defect and the depth under different load amounts were discussed for Digital Shearography in [25], also, the Shearography was combined with digital speckle pattern interferometry (DSPI) to locate defects depth with error less than 5% [26].

However, excessive rigid body displacement that usually occurred with large structure's deformation may cause the speckle of Shearography to lose correlation and greatly reduces the visibility of interferometric fringes. As an alternative method, DIC or FPP (Fringe Projection Profilometry) can be employed to measure the displacement at the level of micro- and millimeter or even above. In our previous work [21] the FPP method can capture the minimum displacement of 10 μm. A DIC-assisted FPP for high-speed 3D (three dimensional) shape, displacement and deformation measurement was presented in [22], where accurate point tracking of 3D shape data leads to precise displacement and deformation measurement with scale of millimeter.

This paper is organized as follows: firstly, a meso-mechanical model of a graded wrinkle is set up and the effective stiffness is derived, which is implanted in finite element simulation to predict the out-of-plane displacement. The trans-scale displacement response is found, which ranges from $10^2$ nm to $10^6$ nm under a series loading conditions. To match the magnitude of displacement, two different methods including Shearography and FPP are presented with an algorithm of displacement extraction, respectively. The experimental results show that Shearography has a higher sensitivity and is suitable for displacement within the scale of nanometer, however FPP has an advantage in the millimeter magnitude. Also, the measurement accuracy of two methods is compared.

## 2 Modeling of wrinkle defect

### 2.1 Wrinkle geometry

The geometry of wrinkle is usually descripted as a uniform or a graded type, for the graded wrinkle (see Fig.3), the amplitude of the waviness is assumed to decay linearly from the maximum at the midsurface to zero on the outer surface [27].

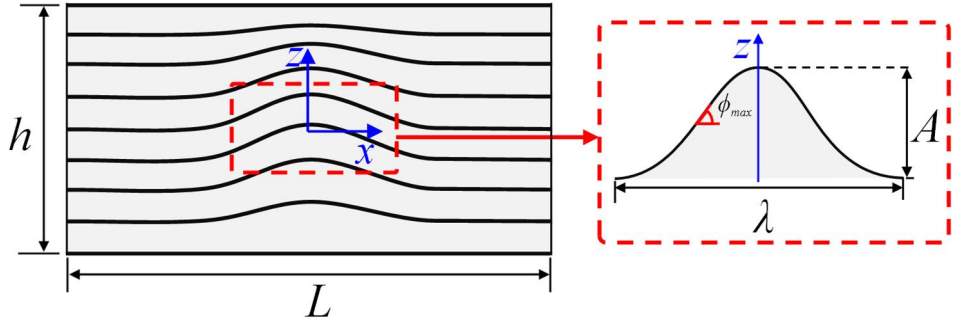

Fig. 3 Geometry of graded wrinkle form

A cosine function is used to descript the geometry of graded wrinkle [28]:

$$W(x,z) = A\left(\frac{h-2|z|}{h}\right)\cos\left(\frac{2\pi x}{\lambda}\right), \quad |x| \leq \lambda/2, \quad |z| \leq h/2 \quad (1)$$

where $x$ is the direction parallel to the fibers without corrugation and $z$ is the direction of thickness. $h$ is the total height of laminate. $A$ and $\lambda$ denotes the wave amplitude and wavelength, respectively.

The ratio of $A/\lambda$ can be used to indicate the degree of severity of wrinkle defect. According to Eq. (1) when $x=\pm\lambda/4$, the misalignment angle reaches maximum.

Table 2 gives the range of maximum misalignment angles and the corresponding ratio of $A/\lambda$ that used in this paper.

Table 2 Range of maximum misalignment angles and ratio of $A/\lambda$

| $A/\lambda$ | 0.10 | 0.15 | 0.20 | 0.25 | 0.30 | 0.35 | 0.40 | 0.45 | 0.50 |
|---|---|---|---|---|---|---|---|---|---|
| $\phi_{max}/°$ | 32.14 | 43.30 | 51.49 | 57.52 | 62.05 | 65.55 | 68.30 | 70.52 | 72.34 |

### 2.2 Meso-mechanical modeling

The representative volume element (RVE) containing a graded wrinkle is considered.

Based on the homogenization method [29], the effective stiffness of such RVE is derived. Fig.4 shows the homogenization process, firstly the RVE is divided into some stripes along horizontal direction. $[C_{ij}]$ is the stiffness of each ply in one strip, which can be transformed via $[\overline{C}_{ij}]$ that denotes the stiffness matrix in the principal coordinate of unidirectional prepreg,

$$\begin{aligned}[C_{ij}] &= [T_\phi]^{-1}[\tilde{C}_{ij}][T_\phi]^{-T}\\ [\tilde{C}_{ij}] &= [T_\theta]^{-1}[\overline{C}_{ij}][T_\theta]^{-T}\end{aligned} \quad (2)$$

where $\theta$ is in-plane orientation angle of fibers and $\phi$ is out-of-plane misalignment angle. $T_\theta$ and $T_\phi$ indicates the transform matrix corresponding to $\theta$ and $\phi$, respectively.

$$[T_\theta] = \begin{bmatrix} \cos^2\theta & \sin^2\theta & 0 & 0 & 0 & 2\sin\theta\cos\theta \\ \sin^2\theta & \cos^2\theta & 0 & 0 & 0 & -2\sin\theta\cos\theta \\ 0 & 0 & 1 & 0 & 0 & 0 \\ 0 & 0 & 0 & \cos\theta & -\sin\theta & 0 \\ 0 & 0 & 0 & \sin\theta & \cos\theta & 0 \\ -\sin\theta\cos\theta & \sin\theta\cos\theta & 0 & 0 & 0 & \cos^2\theta - \sin^2\theta \end{bmatrix} \quad (3)$$

$$[T_\phi] = \begin{bmatrix} \cos^2\phi & 0 & \sin^2\phi & 0 & 2\sin\phi\cos\phi & 0 \\ 0 & 1 & 0 & 0 & 0 & 0 \\ \sin^2\phi & 0 & \cos^2\phi & 0 & -2\sin\phi\cos\phi & 0 \\ 0 & 0 & 0 & \cos\phi & 0 & -\sin\phi \\ -\sin\phi\cos\phi & 0 & \sin\phi\cos\phi & 0 & \cos^2\phi - \sin^2\phi & 0 \\ 0 & 0 & 0 & \sin\phi & 0 & \cos\phi \end{bmatrix} \quad (4)$$

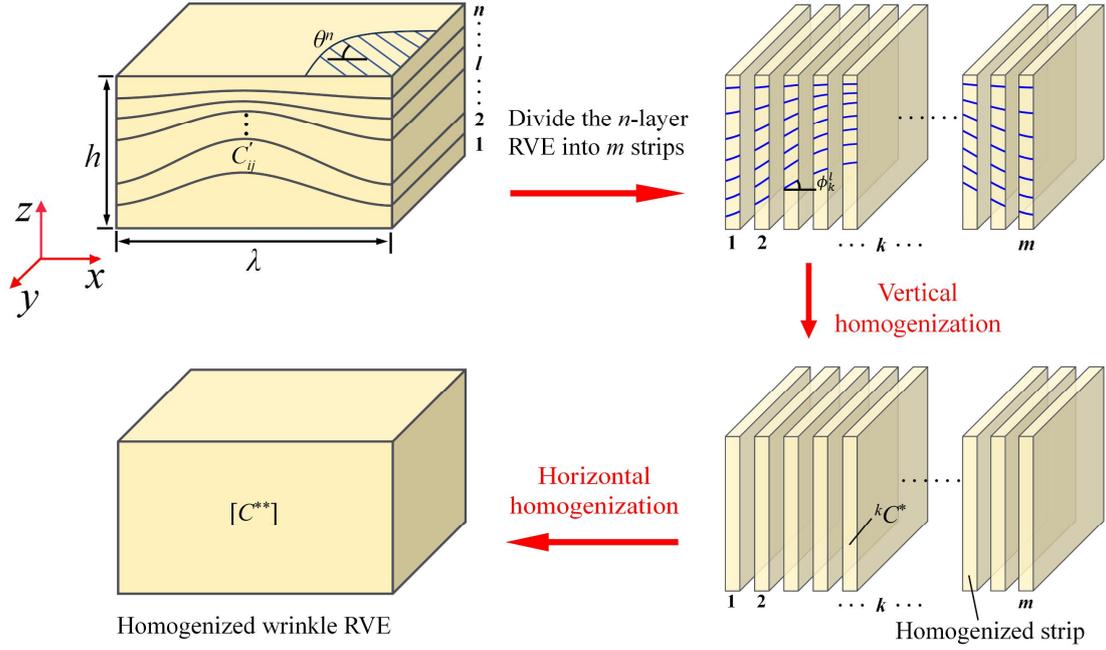

Fig.4 Meso-mechanical modeling of wrinkle

During the vertical homogenization, the out-of-plane stress components $\sigma_{zz}$, $\sigma_{yz}$, $\sigma_{zx}$ and in-plane strain components $\varepsilon_{xx}$, $\varepsilon_{yy}$, $\varepsilon_{xy}$ are assumed to be uniform throughout the thickness, therefore we have $\sigma_{zz}=\sigma_{zz}^*$, $\sigma_{yz}=\sigma_{yz}^*$, $\sigma_{zx}=\sigma_{zx}^*$ and $\varepsilon_{xx}=\varepsilon_{xx}^*$, $\varepsilon_{yy}=\varepsilon_{yy}^*$, $\varepsilon_{xy}=\varepsilon_{xy}^*$. Here the asterisk denotes the average stress and strain.

Then for each ply in one strip, the constitutive relation can be re-written as:

$$\begin{Bmatrix} \{\sigma_A^*\} \\ \{\sigma_B\} \end{Bmatrix} = \begin{bmatrix} [C_{aa}] & [C_{ab}] \\ [C_{ab}]^T & [C_{bb}] \end{bmatrix} \begin{Bmatrix} \{\varepsilon_A\} \\ \{\varepsilon_B^*\} \end{Bmatrix} \quad (5)$$

where:

$$\{\sigma_A^*\} = \{\sigma_{zz}^* \quad \sigma_{yz}^* \quad \sigma_{zx}^*\}^T \quad \{\sigma_B\} = \{\sigma_{xx} \quad \sigma_{yy} \quad \sigma_{xy}\}^T$$
$$\{\varepsilon_A\} = \{\varepsilon_{zz} \quad 2\varepsilon_{yz} \quad 2\varepsilon_{zx}\}^T \quad \{\varepsilon_B^*\} = \{\varepsilon_{xx}^* \quad \varepsilon_{yy}^* \quad 2\varepsilon_{xy}^*\}^T \quad (6)$$

$$[C_{aa}] = \begin{bmatrix} C_{33} & C_{34} & C_{35} \\ C_{43} & C_{44} & C_{45} \\ C_{53} & C_{54} & C_{55} \end{bmatrix} \quad [C_{ab}] = \begin{bmatrix} C_{13} & C_{23} & C_{36} \\ C_{14} & C_{24} & C_{46} \\ C_{15} & C_{25} & C_{56} \end{bmatrix} \quad [C_{bb}] = \begin{bmatrix} C_{11} & C_{12} & C_{16} \\ C_{21} & C_{22} & C_{26} \\ C_{16} & C_{26} & C_{66} \end{bmatrix} \quad (7)$$

Eq. (5) can be transformed in the term of $\{\varepsilon_B^*\}$ and $\{\sigma_A^*\}$:

$$\begin{Bmatrix} \{\varepsilon_A\} \\ \{\sigma_B\} \end{Bmatrix} = \begin{bmatrix} [C_{aa}]^{-1} & -[C_{aa}]^{-1}[C_{ab}] \\ [C_{ab}]^T[C_{aa}]^{-1} & -[C_{ab}]^T[C_{aa}]^{-1}[C_{ab}]+[C_{bb}] \end{bmatrix} \begin{Bmatrix} \{\sigma_A^*\} \\ \{\varepsilon_B^*\} \end{Bmatrix} \quad (8)$$

$\{\varepsilon_A^*\}$ and $\{\sigma_B^*\}$, as the average of $\{\varepsilon_A\}$ and $\{\sigma_B\}$, can be obtained:

$$\begin{Bmatrix} \{\varepsilon_A^*\} \\ \{\sigma_B^*\} \end{Bmatrix} = \begin{Bmatrix} [C_A^*] & -[C_B^*] \\ [C_B^*]^T & [C_D^*] \end{Bmatrix} \begin{Bmatrix} \{\sigma_A^*\} \\ \{\varepsilon_B^*\} \end{Bmatrix} \tag{9}$$

where:

$$\begin{cases} [C_A^*] = \dfrac{1}{h} \sum_{k=1}^{n} \int_{z^{k-1}}^{z^k} [C_{aa}]^{-1} \, dz \\ [C_B^*] = \dfrac{1}{h} \sum_{k=1}^{n} \int_{z^{k-1}}^{z^k} [C_{aa}]^{-1} [C_{ab}] \, dz \\ [C_D^*] = \dfrac{1}{h} \sum_{k=1}^{n} \int_{z^{k-1}}^{z^k} \left( -[C_{ab}]^T [C_{aa}]^{-1} [C_{ab}] + [C_{bb}] \right) dz \end{cases} \tag{10}$$

$z^k$ and $z^{k-1}$ in Eq. (10) indicates the top and bottom coordinate of each ply, $n$ is the total number of plies.

According to Eq. (9), the constitutive relationship $[C_{ij}^*]$ of each strip via above vertical homogenization process can be obtained:

$$\begin{Bmatrix} \{\sigma_A^*\} \\ \{\sigma_B^*\} \end{Bmatrix} = \begin{Bmatrix} [C_A^*]^{-1} & [C_A^*]^{-1}[C_B^*] \\ [C_B^*]^T [C_A^*]^{-1} & [C_B^*][C_A^*]^{-1}[C_B^*] + [C_D^*] \end{Bmatrix} \begin{Bmatrix} \{\varepsilon_A^*\} \\ \{\varepsilon_B^*\} \end{Bmatrix} \tag{11}$$

In the next horizontal homogenization, it is assumed $\sigma_{xx}^* = \sigma_{xx}^{**}$, $\sigma_{zx}^* = \sigma_{zx}^{**}$, $\sigma_{xy}^* = \sigma_{xy}^{**}$, $\varepsilon_{yy}^* = \varepsilon_{yy}^{**}$, $\varepsilon_{zz}^* = \varepsilon_{zz}^{**}$, $\varepsilon_{yz}^* = \varepsilon_{yz}^{**}$. The superscript ** indicates the twice homogenization result.

$$\begin{Bmatrix} \{\sigma_E^{**}\} \\ \{\sigma_F^*\} \end{Bmatrix} = \begin{Bmatrix} [C_{ee}^*] & [C_{ef}^*] \\ [C_{ef}^*]^T & [C_{ff}^*] \end{Bmatrix} \begin{Bmatrix} \{\varepsilon_E^*\} \\ \{\varepsilon_F^{**}\} \end{Bmatrix} \tag{12}$$

where:

$$\{\sigma_E^{**}\} = \{\sigma_{xx}^{**} \quad \sigma_{zx}^{**} \quad \sigma_{xy}^{**}\}^T \quad \{\sigma_F^*\} = \{\sigma_{yy}^* \quad \sigma_{zz}^* \quad \sigma_{yz}^*\}^T$$
$$\{\varepsilon_E^*\} = \{\varepsilon_{xx}^* \quad 2\varepsilon_{zx}^* \quad 2\varepsilon_{xy}^*\}^T \quad \{\varepsilon_F^{**}\} = \{\varepsilon_{yy}^{**} \quad \varepsilon_{zz}^{**} \quad 2\varepsilon_{yz}^{**}\}^T \tag{13}$$

$$[C_{ee}^*] = \begin{bmatrix} C_{11}^* & C_{15}^* & C_{16}^* \\ C_{15}^* & C_{55}^* & C_{56}^* \\ C_{16}^* & C_{56}^* & C_{66}^* \end{bmatrix} \quad [C_{ef}^*] = \begin{bmatrix} C_{12}^* & C_{13}^* & C_{14}^* \\ C_{25}^* & C_{35}^* & C_{45}^* \\ C_{26}^* & C_{36}^* & C_{46}^* \end{bmatrix} \quad [C_{ff}^*] = \begin{bmatrix} C_{22}^* & C_{23}^* & C_{24}^* \\ C_{23}^* & C_{33}^* & C_{34}^* \\ C_{24}^* & C_{34}^* & C_{44}^* \end{bmatrix} \tag{14}$$

Eq. (12) is rewritten as:

$$\begin{Bmatrix} \{\varepsilon_E^*\} \\ \{\sigma_F^*\} \end{Bmatrix} = \begin{Bmatrix} [C_{ee}^*]^{-1} & -[C_{ee}^*]^{-1}[C_{ef}^*] \\ \left[[C_{ee}^*]^{-1}[C_{ef}^*]\right]^T & -[C_{ef}^*]^T[C_{ee}^*]^{-1}[C_{ef}^*] + [C_{ff}^*] \end{Bmatrix} \begin{Bmatrix} \{\sigma_E^{**}\} \\ \{\varepsilon_F^{**}\} \end{Bmatrix} \tag{15}$$

$\{\varepsilon_E^{**}\}$ and $\{\sigma_F^{**}\}$, as the average of $\{\varepsilon_E^*\}$ and $\{\sigma_F^*\}$, can be obtained:

$$\begin{Bmatrix} \{\varepsilon_E^{**}\} \\ \{\sigma_F^{**}\} \end{Bmatrix} = \begin{bmatrix} [C_E^{**}]^{-1} & -[C_F^{**}] \\ [C_F^{**}]^T & [C_H^{**}] \end{bmatrix} \begin{Bmatrix} \{\sigma_E^{**}\} \\ \{\varepsilon_F^{**}\} \end{Bmatrix} \quad (16)$$

where:

$$\begin{cases} [C_E^{**}] = \dfrac{1}{\lambda} \int_{-\lambda/2}^{\lambda/2} [C_{ee}^*]^{-1} \, \mathrm{d}x \\ [C_F^{**}] = \dfrac{1}{\lambda} \int_{-\lambda/2}^{\lambda/2} [C_{ee}^*]^{-1} [C_{ef}^*] \, \mathrm{d}x \\ [C_H^{**}] = \dfrac{1}{\lambda} \int_{-\lambda/2}^{\lambda/2} \left( -[C_{ef}^*]^T [C_{ee}^*]^{-1} [C_{ef}^*] + [C_{ff}^*] \right) \mathrm{d}x \end{cases} \quad (17)$$

According to Eq. (16) the effective stiffness $[C_{ij}^{**}]$ of the RVE containing a graded wrinkle can be obtained:

$$\begin{Bmatrix} \{\sigma_E^{**}\} \\ \{\sigma_F^{**}\} \end{Bmatrix} = \begin{Bmatrix} [C_E^*]^{-1} & [C_E^*]^{-1}[C_F^*] \\ [C_F^*]^T[C_E^*]^{-1} & [C_F^*]^T[C_E^*]^{-1}[C_F^*] + [C_H^*] \end{Bmatrix} \begin{Bmatrix} \{\varepsilon_E^{**}\} \\ \{\varepsilon_F^{**}\} \end{Bmatrix} \quad (18)$$

## 3 Finite element prediction

The finite element analysis (FEA) is used to explore the displacement response with variation of structural parameters. Based on our previous study [30], the out-of-plane displacement of a wrinkled laminate has a significant response to the tension load rather than torque, bending or shearing loads. It can be an indicator to locate the defects and gives the assessment of the degradation of the material.

Fig.5 shows the finite element model of a wrinkle defect embedded in the middle of the specimen, and a certain range of thickness is considered. Table 3 lists the structural parameters and wrinkle sizes.

The effective stiffness derived from Section 2 is employed in the FEA to predict the maximum out-of-plane displacement response. The mechanical properties of carbon/epoxy prepreg can be seen in Table 4.

Table 3 Structural parameters of the laminates with wrinkle defects

| No. | Lay-up | plies | Winkle parameters | | |
|---|---|---|---|---|---|
| | | | Wave length $\lambda$ /mm | Wave amplitude $A$ /mm | Ratio $A/\lambda$ |
| ① | $[0/90]_{2s}$ | 8 | 5 | 0.5 | 0.1 |
| | | | | 0.75 | 0.15 |
| ② | $[0/90]_{4s}$ | 16 | 5 | 0.5 | 0.1 |
| | | | | 1 | 0.2 |
| ③ | $[0/90/\pm45/0]_{3s}$ | 30 | 5 | 1 | 0.2 |
| | | | | 1.75 | 0.25 |
| | | | | 2.5 | 0.5 |

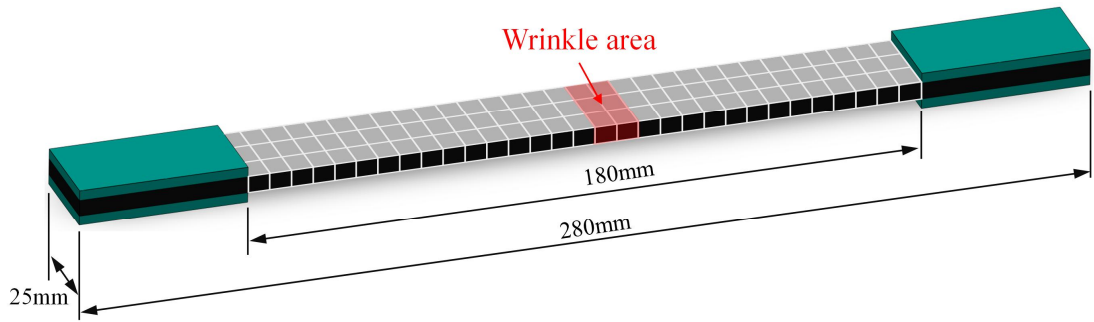

Fig. 5 Finite element model

Table 4 Material properties of carbon/epoxy

| $E_{11}$/GPa | $(E_{22}/E_{33})$/GPa | $G_{23}$/GPa | $G_{31}$/GPa | $G_{12}$/GPa | $v_{21}$ | $v_{32}$ | $v_{31}$ |
|---|---|---|---|---|---|---|---|
| 133.3 | 9.09 | 3.16 | 7.24 | 7.23 | 0.261 | 0.436 | 0.261 |

Fig. 6 shows the tension results of FEA, where the out-of-plane displacement becomes obvious and the maximum displacement occurs in the wrinkled area. However, in the laminate without wrinkle defect the off-plane deformation is uniform.

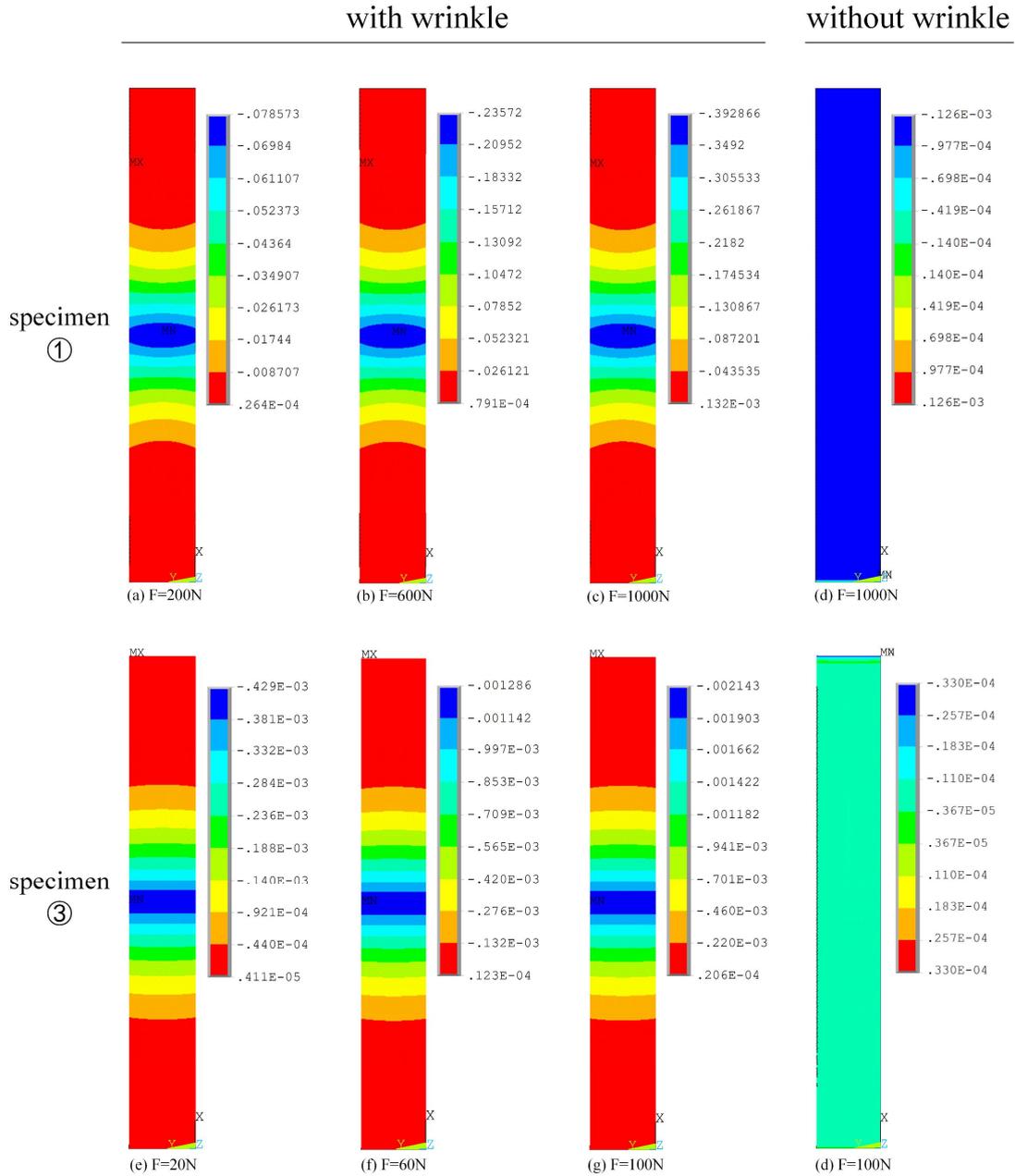

Fig. 6 Finite element results of out-of-plane displacement (units: mm)

The right of Fig.7 shows the out-of-plane displacement exhibiting a significant trans-scale effect, where logarithmic coordinate is used to display this effect. The displacement varies from $10^2$nm to $10^6$nm with different wrinkle defects and structural parameters, which means different approaches should be used to measure the displacement response. Usually, Shearography covers the measurement range within $10^3$nm, and DIC can provide a micrometer scale measurement. For the displacement reaches millimeter-scale, we proposed FPP method combining with displacement extraction algorithm.

Experimental investigation is conducted in the next section and two methods including Shearography and FPP are employed to reveal the wrinkle's response, the sensitivity and accuracy of two methods are compared, which is helpful to the future application in industrial non-destructive test.

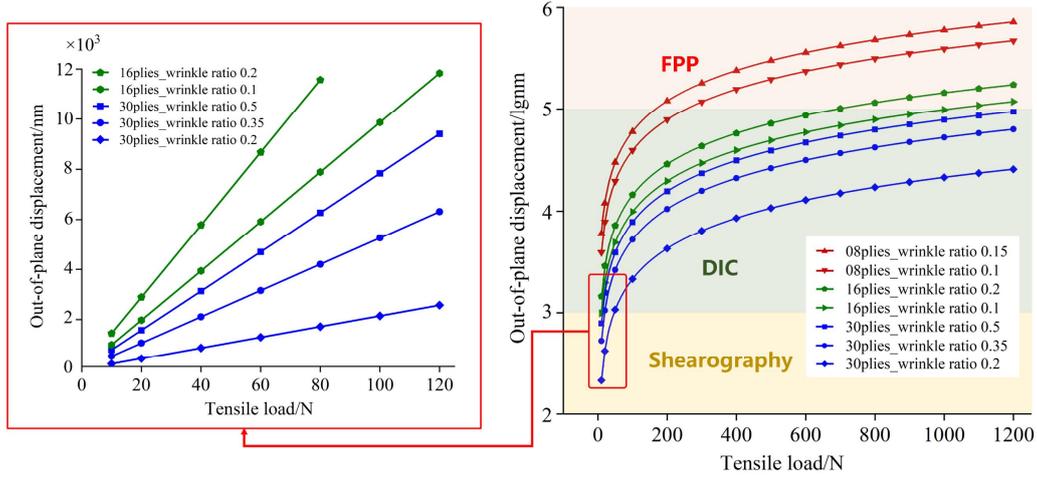

Fig. 7 Out-of-plane displacement of wrinkled laminates

## 4 Specimen prepare

The specimens were made as following steps [31]: firstly, the prepreg was cut with dimensions of 300 mm × 300 mm and laid up with different number of layers, where a steel rod was placed at the bottom layer of the prepreg. Before the heat curing process, the steel rod was removed and the epoxy resin in its liquid state will converge into the void caused by the steel rod during the curing. Then the finished laminate was strengthened at two ends and to be cut into specimens with a length of 280 mm and a width of 25 mm. Fig.8 illustrates the profile of such specimens.

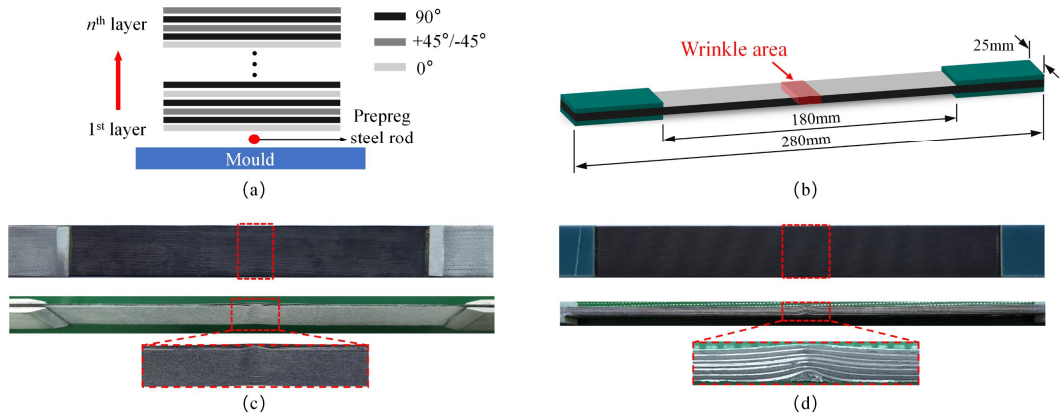

Fig. 8 Specimens with wrinkle defects

The detail wrinkle parameters can be seen in Table 5.

Table 5 Wrinkle parameters

| No. | Lay-up | plies | Winkle parameters | | |
|---|---|---|---|---|---|
| | | | Wave length $\lambda$ /mm | Wave amplitude $A$ /mm | Ratio $A/\lambda$ |
| I | $[0/90]_{2s}$ | 8 | 6.6 | 1.2 | 0.18 |
| II | $[0]_{30}$ | 30 | 8.3 | 1.0 | 0.12 |

## 5 Shearography measurement
### 5.1 Experimental apparatus

Fig. 9 shows the schematic set-up in case of Shearographic measurement. The specimen is fixed in the Instron test machine and is illuminated by He-Ne lasers. Due to the diffusely scattering surface of specimen, a portion of the light is reflected to the CCD camara and thus is shown on the focal plane by the lens system. The light passes a Michelson interferometer, by tilting one of the mirrors of the Michelson interferometer the so-called shear mirror, two different points of the specimen's surface can be superimposed on one point of the focal plane. Therefore, the corresponding beam paths interfere on the focal plane.

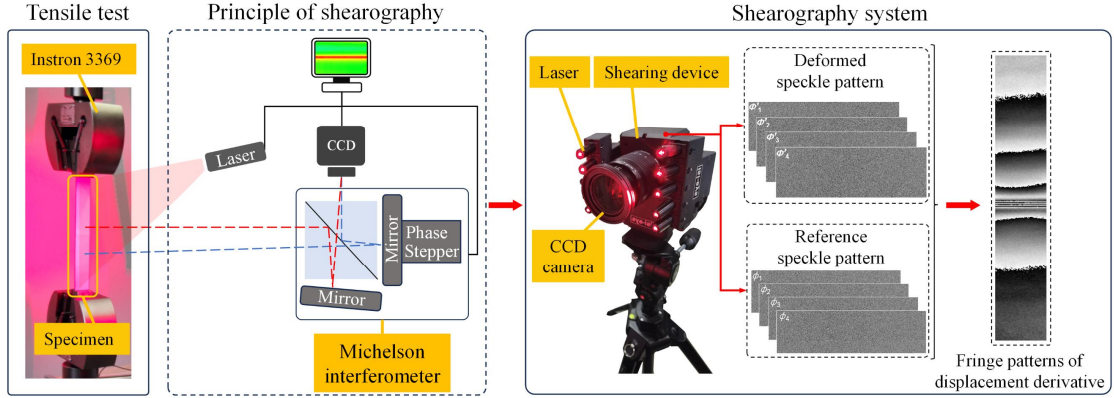

Fig. 9 Experimental device with shearography system

For the ISI Shearography system that we employed in this experiment, the angle between the illumination light and the observation direction is approximately zero, thus the relative phase difference $\Delta\Phi_y$ and the out-of-plane displacement derivative $\partial w/\partial y$ has the relation:

$$\Delta\Phi_y = \frac{4\pi}{\lambda_L}\frac{\partial w}{\partial y}\Delta y \qquad (19)$$

where $\lambda_L$ is the wavelength of He-Ne laser, $\Delta y$ is the shearing amount that controls the sensitivity of the measurement.

### 5.2 Experimental Results

Fig. 10 is the measured wrapped phase map with a phase modulo $2\pi$, corresponding to the out-of-plane displacement derivative of specimen I and II, respectively, where the spacing of fringes becomes narrow with the increment of tension loads.

Via phase unwrapping process, the wrapped phase fringe discontinuities can be removed to obtain a continuous distribution of out-of-displacement derivative. An arbitrary point in the bottom of the specimen is taken as a reference point with zero displacement during the definite integral, the out-of-plane displacement can be calculated. Fig.11 shows the unwrapping and integral process. The obtained maximum displacement is from 350nm to 1500nm (specimen I), and 500nm to 2400nm (specimen II), respectively.

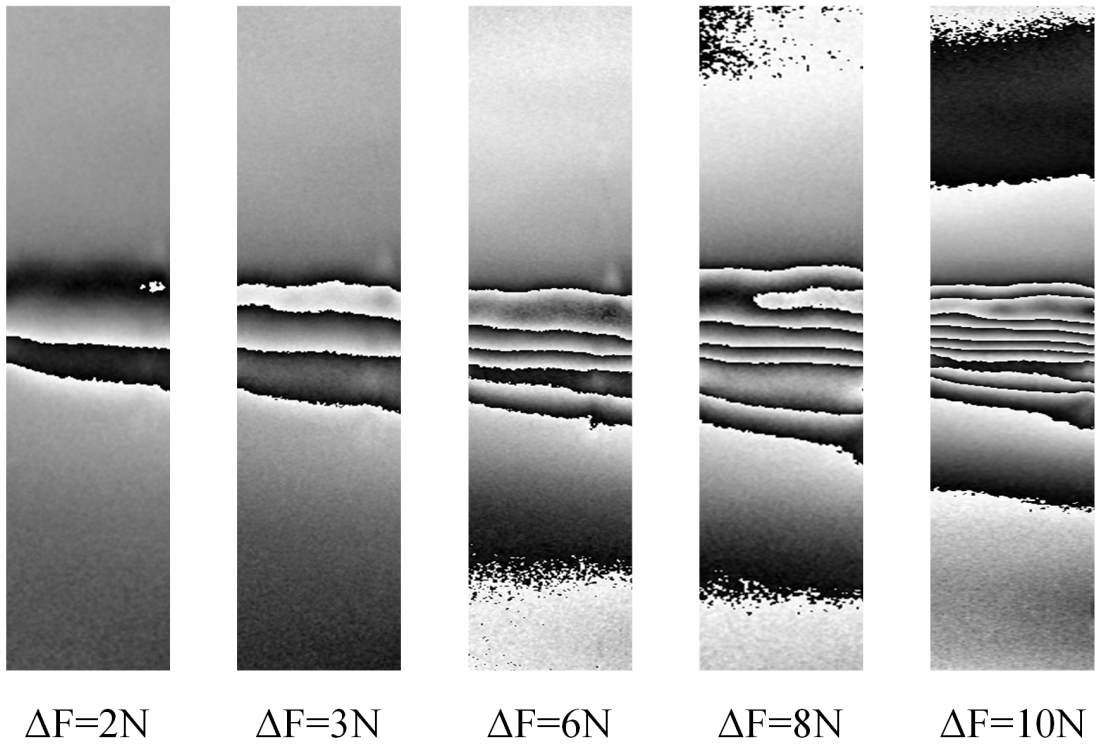

ΔF=2N　　　ΔF=3N　　　ΔF=6N　　　ΔF=8N　　　ΔF=10N

(a) Specimen I

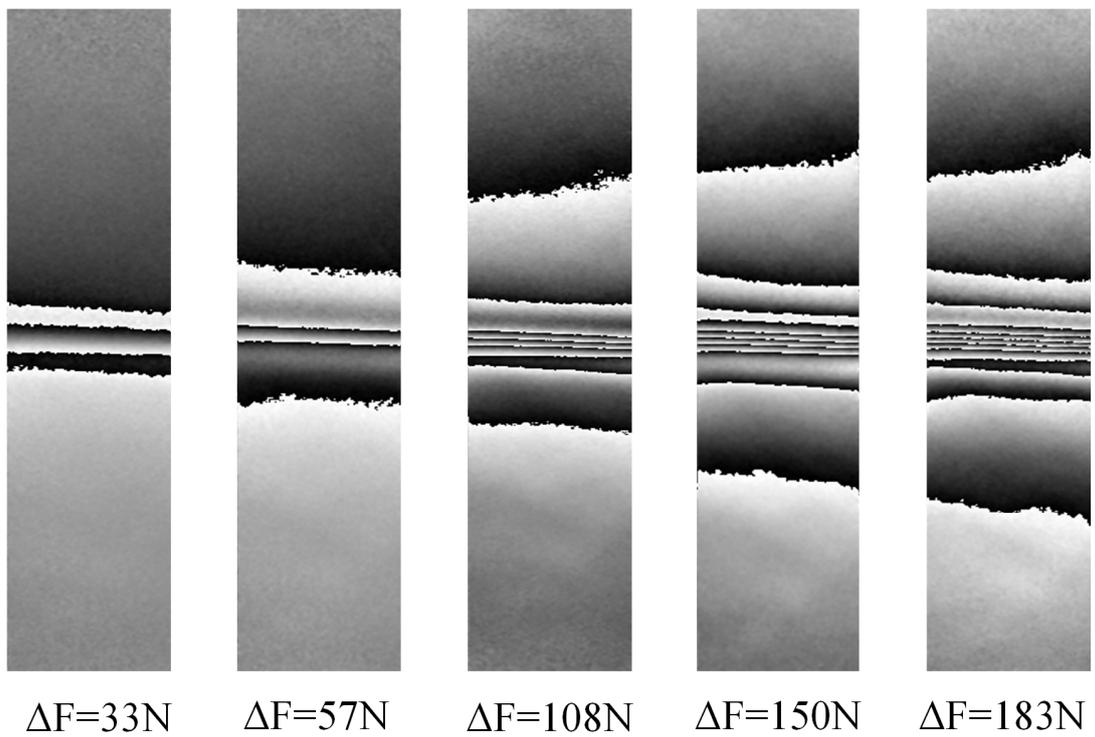

ΔF=33N　　　ΔF=57N　　　ΔF=108N　　　ΔF=150N　　　ΔF=183N

(b) Spcimen II

Fig. 10 Fringe patterns of specimen I and II during tension test

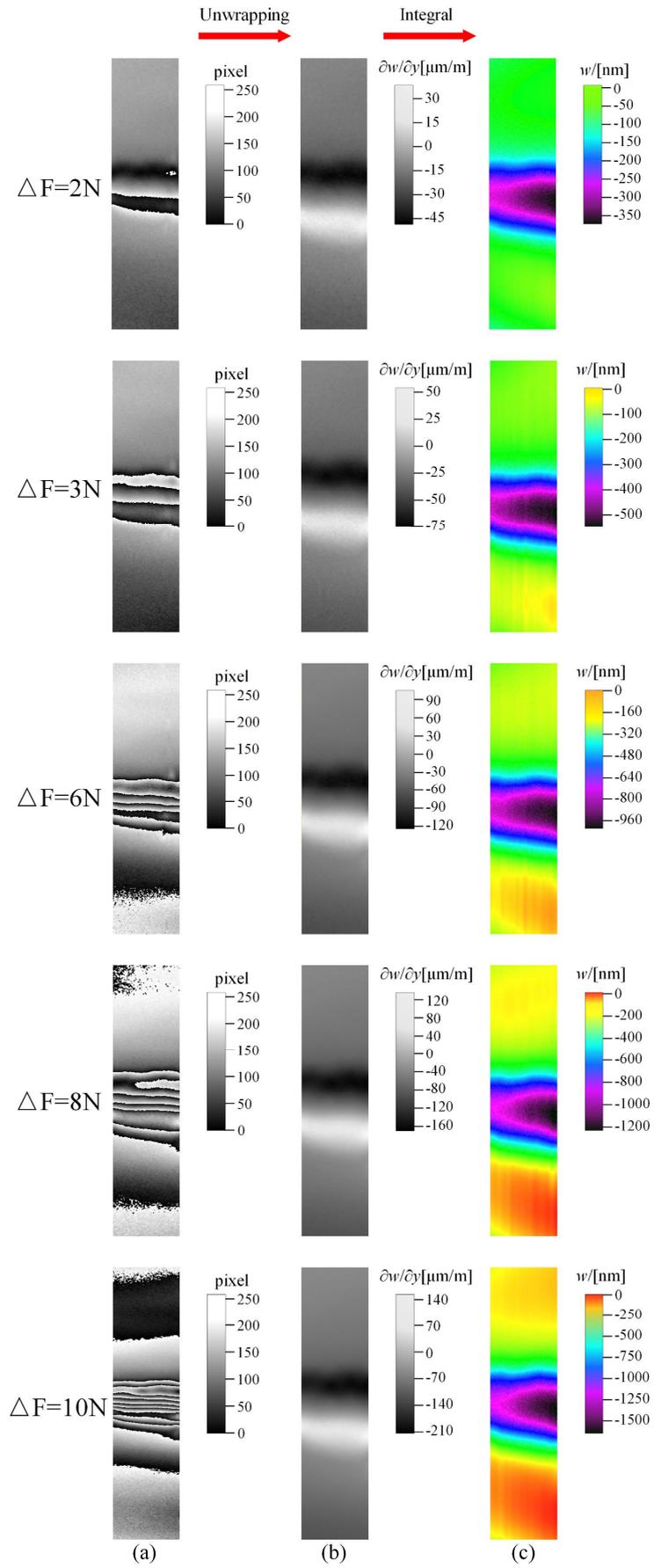

△F=2N

△F=3N

△F=6N

△F=8N

△F=10N

(a)　　　(b)　　　(c)

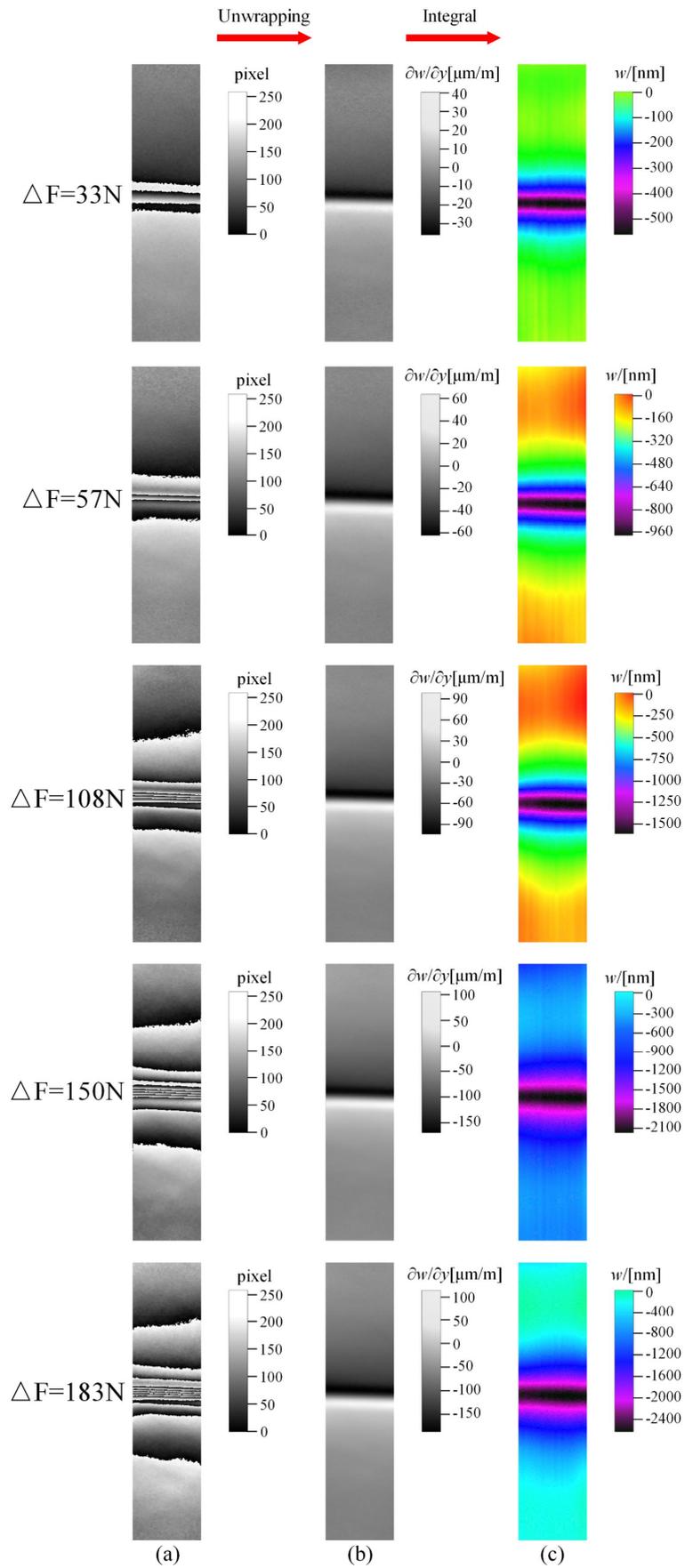

Fig. 11 Unwrapping and integral process to obtain out-of-plane displacement

## 6 FPP measurement
### 6.1 Experimental device

The FPP system includes two CCD cameras and a fringe projector. The pseudo-sinusoidal fringes are projected on the surface of the specimen, which is fixed in the Instron test machine and the cameras capture the deformed patterns as shown in Fig.12.

At every load step, the depth information of the specimens that encoded in the deformed fringe patterns is acquired from three wrapped phases with $2\pi/3$ phase-shift via phase unwrapping and shape reconstruction, which is represented as a 3D point cloud data.

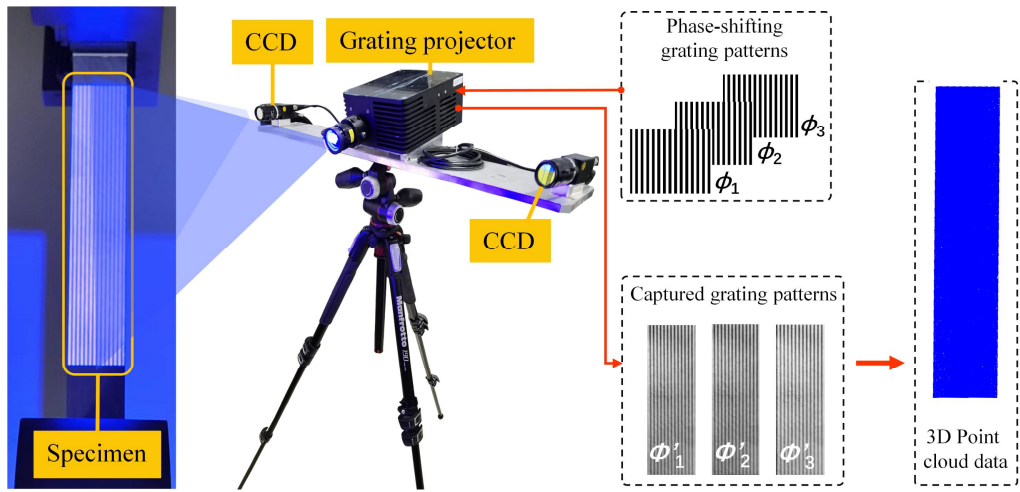

Fig.12 Schematic diagram of 3D shape measuring by FPP

### 6.2 Displacement Extraction

With the increasing tensile load, the topography of specimen was recorded by dense and accurate 3D point coordinates. Theoretically, the out-of-plane displacement can be acquired from two groups of point clouds, however, it is impractical due to the lack of point correspondence. Therefore, we proposed an algorithm to regulate the 3D coordinates. Fig. 13 shows that the cubic spline interpolation was used to obtain the value of height on the same mesh grid in plane X-Y, then the subtraction of two new groups of point clouds gives the accurate out-of-plane displacement. This method works well because the in-plane displacement is far smaller than the out-of-plane displacement.

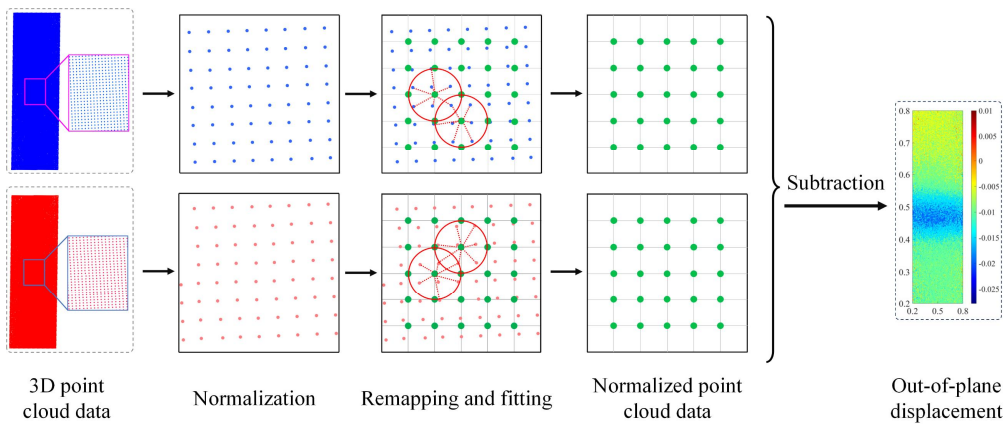

Fig.13 Displacement extraction algorithm

## 6.3 Experimental Result

Fig.14 shows the contours of the depths of the specimen I under different tensile loads. Combining with the displacement extraction algorithm, the out-of-plane displacement can be obtained in Fig.15. It can be found that the wrinkled area has a significant augment of out-of-plane displacement compared the field with perfect material properties. The out-of-plane displacement increases from 0.065mm to 0.280mm with the tensile force increases from 200N to 1000N.

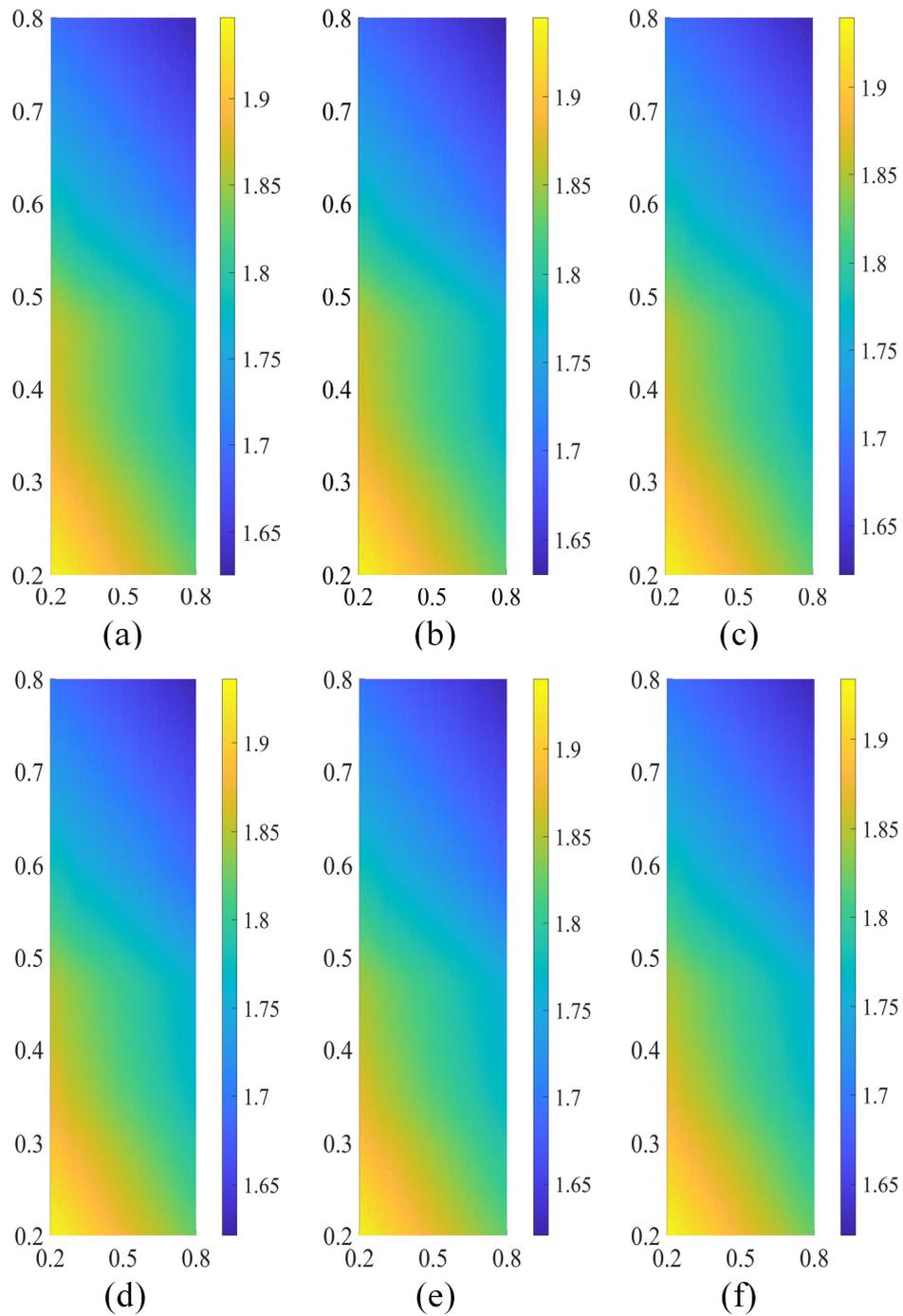

Fig.14 Depth of specimen A2 (unit: cm): (a) F=0N, (b) F=200N, (c) F=400N, (d) F=600N, (e) F=800N, (f) F=1000N.

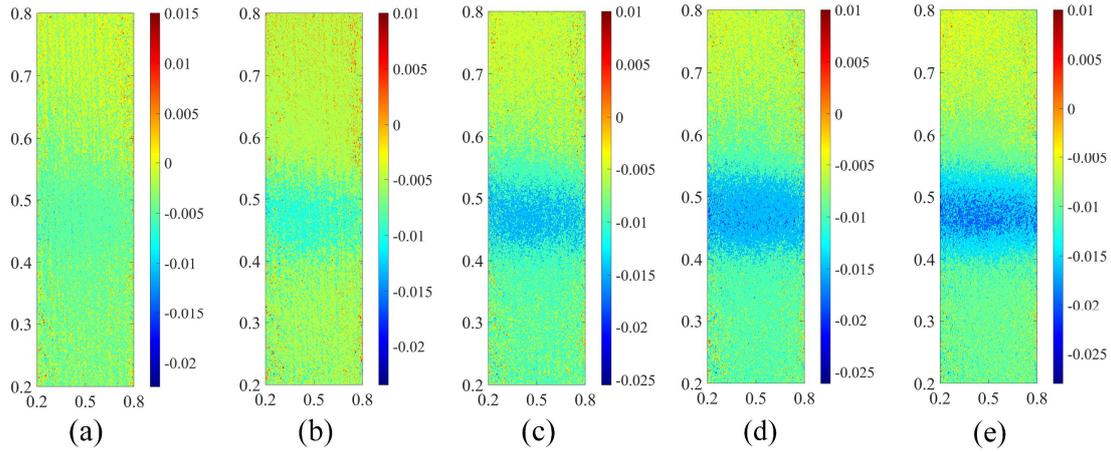

Fig.15 Out-of-plane displacement of specimen I (unit:cm): (a) ΔF=200N, (b) ΔF=400N, (c) ΔF=600N, (d) ΔF=800N, (e) ΔF=1000N.

## 7 Methods Comparison

For the specimen I with a lower thickness, two methods were both employed to measure the out-of-plane displacement. Fig.16 gives the comparison of Shearography result and FEA result along the scale line. The measured maximum displacement ranges from 375nm to 1631nm, according to the tensile force from 2N to 10N, and Table 6 gives the error analysis at every load step.

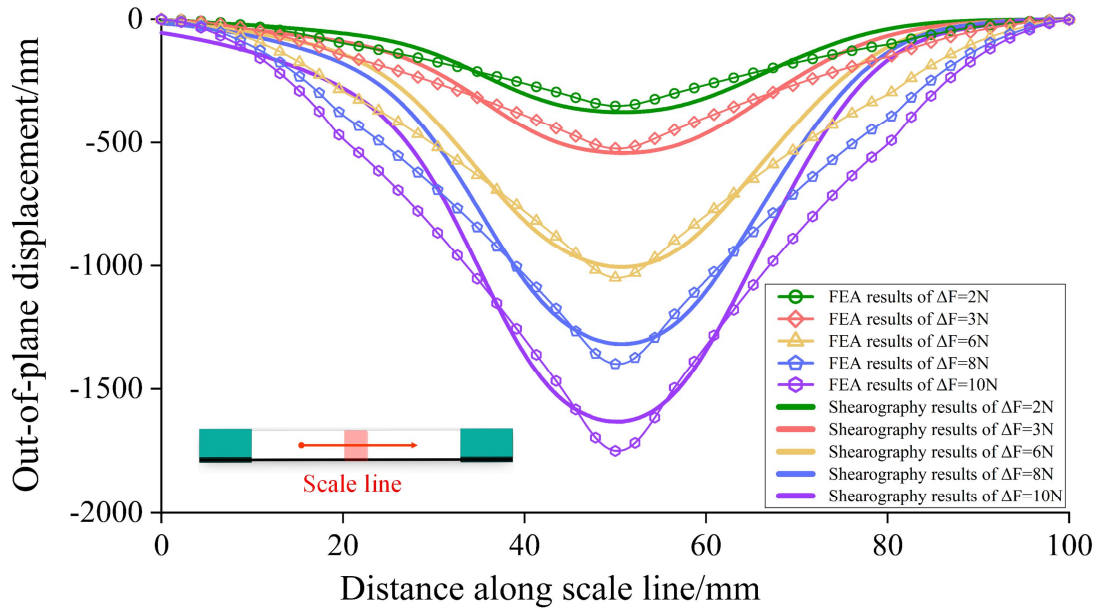

Fig.16 Comparison of shearography measurement and FEA

Table.6 Error of shearography results and FEA results

| Specimen | Load/N | Maximum out-of-plane displacement | | Error |
| --- | --- | --- | --- | --- |
| | | Shearography results/nm | FEA results/nm | |
| I | 2 | -375 | -350 | 6.7% |
| | 3 | -542 | -524 | 3.3% |
| | 6 | -1006 | -1048 | 4.2% |
| | 8 | -1317 | -1398 | 6.2% |
| | 10 | -1631 | -1748 | 7.1% |

The left part of Fig. 17 shows the maximum displacement comparison of FPP results and that of FEA, according to the load from 200N to 1000N, and the displacement distribution along the scale line with tension force 600N is compared in the right part of Fig.17. Also, Table 7 gives the error analysis.

It is obviously Shearography has a higher sensitivity than FPP, the former can capture the displacement responses with a very small tension force, while the latter requires more than 100 times the force. However, it does not mean that Shearography always has a greater advantage. FPP also has its own superiority when large deformations need to be detected. Here we provide a quantitative test data for two methods, and researchers can choose appropriate one to meet their own demand.

The error range between Shearography and FEA results is 3.3%~7.1%, and that is 4.1%~15.6% between FPP and FEA results. Such result indicates the two methods have a close accuracy in displacement measurement.

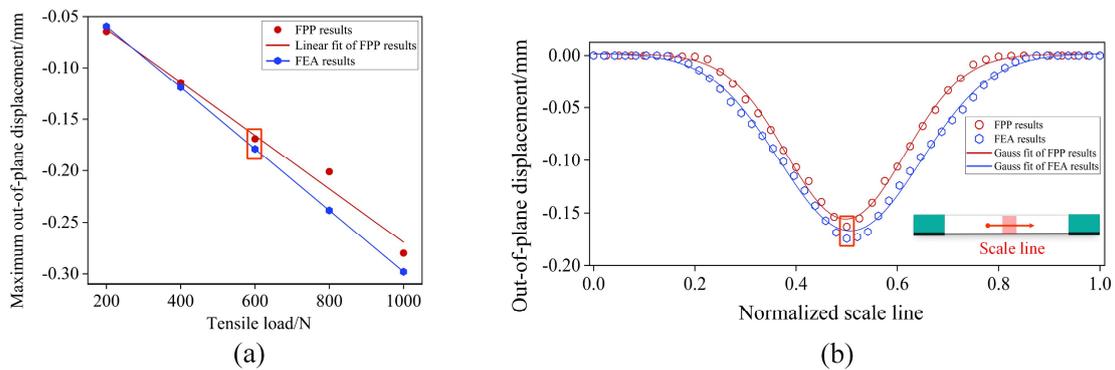

Fig.17 (a) Maximum out-of-plane displacement of specimen I obtained by FPP and FEA; (b) Out-of-plane displacement of specimen I under 600 N

Table 7 Error between FPP results and FEA results

| Specimen | Load/N | Maximum out-of-plane displacement | | Error |
|---|---|---|---|---|
| | | FPP results/mm | FEA results/mm | |
| I | 200 | -0.0646 | -0.0596 | 8.4% |
| | 400 | -0.1142 | -0.1191 | 4.1% |
| | 600 | -0.169 | -0.1788 | 5.5% |
| | 800 | -0.2011 | -0.2384 | 15.6% |
| | 1000 | -0.280 | -0.298 | 6.0% |

## 8 Conclusions

Wrinkle defects are found widely existed in wind turbine blades, filament-wound composite pressure vessels and other laminated structures. This paper establishes a meso-mechanical model of wrinkle defect, obtains the effective stiffness of wrinkled composite material, and which is implanted into FEA to predict the out-of-plane displacement response. The numerical simulation shows that the out-of-plane displacement in the wrinkled area is highly sensitive to the tensile load. Different sizes of wrinkle defects that embedded in specimens causes the displacement ranging from $10^2 \sim 10^6$ nanometers, which poses different requirements for the out-of-plane displacement measurement methods.

There are three methods available for measuring the out-of-plane displacement. Among them, ESPI is suitable for the displacement within the level at $10^3$ nanometers, 3D DIC is suitable for the measurement around the micrometer level, and FPP method can be used for the displacement at the millimeter level or even larger. Due to the wide usage of DIC method, this paper mainly focusses on the shearography and FPP methods and the experimental investigation was conducted.

The experimental results show that the sensitivity of shearography is much higher than that of FPP. For the same thin-walled specimen, a tensile force less than 10N is large enough to produce the detectable response by shearography. While with the FPP method, the load needs to be increased to 200N~1000N and the corresponding displacement can therefore be measured by the FPP. However, it does not mean that the shearography is overwhelmingly superior to FPP, because in some industrial scenario, the rigid body displacement may be significant and destroys the visibility of speckle interferometric fringes.

The comparison with the FEA shows that the error of shearography and FPP is 3.3%~7.1% and 4.1%~15.6%, respectively, which indicates the two methods have a similar measurement accuracy.

**Acknowledgement**

This work was supported by the National Key Research & Development Program of China [grant number: 2021YFB4000903] and Basic Public Welfare Research Program of Zhejiang Province [grant number: LGF21E040002].

## References

[1] Zhang J, Lin G, Vaidya U, et al. Past, present and future prospective of global carbon fibre composite developments and applications. Composites Part B: Engineering. 2023; 250:110463.

[2] Yang H, Yang L, Yang Z, et al. Ultrasonic detection methods for mechanical characterization and damage diagnosis of advanced composite materials: A review. Composite Structures. 2023; 324:117554.

[3] Ju X, Xiao J, Wang D, et al. Effect of gaps/overlaps induced waviness on the mechanical properties of automated fiber placement (AFP)-manufactured composite laminate. Mater Res Express. 2022; 9:045305.

[4] Nsengiyumva W, Zhong S, Lin J, et al. Advances, limitations and prospects of nondestructive testing and evaluation of thick composites and sandwich structures: A state-of-the-art review. Composite Structures. 2021; 256:112951.

[5] Shigang A, Daining F, Rujie H, et al. Effect of manufacturing defects on mechanical properties and failure features of 3D orthogonal woven C/C composites. Composites Part B: Engineering. 2015; 71:113–121.

[6] Fu Y, Yao X. A review on manufacturing defects and their detection of fiber reinforced resin matrix composites. Composites Part C: Open Access. 2022;8:100276.

[7] Chen X. Fracture of wind turbine blades in operation—Part I: A comprehensive forensic investigation. Wind Energy. 2018; 21:1046–1063.

[8] Miao X-Y, Chen C-J, Fæster S, et al. Fatigue and post-fatigue static crack characterisation of a wrinkled thick glass fibre laminate in a composite wind turbine blade. International Journal of Fatigue. 2023; 176:107855.

[9] Towsyfyan H, Biguri A, Boardman R, et al. Successes and challenges in non-destructive testing of aircraft composite structures. Chinese Journal of Aeronautics. 2020; 33:771–791.

[10] Garcea SC, Wang Y, Withers PJ. X-ray computed tomography of polymer composites. Composites Science and Technology. 2018; 156:305–319.

[11] Wilhelmsson D, Gutkin R, Edgren F, et al. An experimental study of fibre waviness and its effects on compressive properties of unidirectional NCF composites. Composites Part A: Applied Science and Manufacturing. 2018; 107:665–674.

[12] Calvo JV, Quiñonero-Moya AR, Feito N, et al. Influence of distributed out-of-plane waviness defects on the mechanical behavior of CFRP laminates. Composite Structures. 2023; 323:117498.

[13] Mendonça HG, Mikkelsen LP, Zhang B, et al. Fatigue delaminations in composites for wind turbine blades with artificial wrinkle defects. International Journal of Fatigue. 2023; 175:107822.

[14] Spencer M, Chen X. Static and fatigue cracking of thick carbon/glass hybrid composite laminates with complex wrinkle defects. International Journal of Fatigue. 2023; 177:107963.

[15] Davidson P, Waas AM. The effects of defects on the compressive response of thick carbon composites: An experimental and computational study. Composite Structures. 2017; 176:582-596.

[16] Bloom LD, Wang J, Potter KD. Damage progression and defect sensitivity: An experimental study of representative wrinkles in tension. Composites Part B: Engineering. 2013; 45:449-458.

[17] Hallander P, Sjölander J, Åkermo M. Forming induced wrinkling of composite laminates with

mixed ply material properties; an experimental study. Composites Part A: Applied Science and Manufacturing. 2015; 78:234-245.

[18] Mizukami K, Mizutani Y, Todoroki A, et al. Detection of in-plane and out-of-plane fiber waviness in unidirectional carbon fiber reinforced composites using eddy current testing. Composites Part B: Engineering. 2016; 86:84-94.

[19] Naderi, M.; Ji, M.; Liyanage, S.; Palliyaguru, U.; Soghrati, S.; Iyyer, N.; et al. Experimental and numerical analysis of wrinkles influence on damage mechanisms and strength of L-Shape cross-ply composite beams. Composites Science and Technology. 2022; 223:109420.

[20] Nebe M. In Situ Characterization Methodology for the Design and Analysis of Composite Pressure Vessels [Internet]. Wiesbaden: Springer Fachmedien Wiesbaden; 2022 [cited 2024 Feb 7]. Available from: https://link.springer.com/10.1007/978-3-658-35797-9.

[21] Wen A, Ma L, Shen C, et al. Wrinkle defects investigation on displacement response in fiber-reinforced composites using grating projection measurement. Composite Structures. 2022; 297:115867.

[22] Wu Z, Guo W, Pan B, et al. A DIC-assisted fringe projection profilometry for high-speed 3D shape, displacement and deformation measurement of textured surfaces. Optics and Lasers in Engineering. 2021; 142:106614.

[23] Francis D, Tatam RP, Groves RM. Shearography technology and applications: a review. Meas Sci Technol. 2010; 21:102001.

[24] Zhang L, Chen YF, Liu H, et al. In-situ real-time imaging of subsurface damage evolution in carbon fiber composites with shearography. Composites Communications. 2022; 32:101170.

[25] Guo B, Zheng X, Gerini-Romagnoli M, et al. Digital shearography for NDT: Determination and demonstration of the size and the depth of the smallest detectable defect. NDT & E International. 2023; 139:102927.

[26] Gu G, Pan Y, Qiu C, et al. Improved depth characterization of internal defect using the fusion of shearography and speckle interferometry. Optics & Laser Technology. 2021; 135:106701.

[27] Thalapil J, Sawant S, Tallur S, Banerjee S. Guided wave based localization and severity assessment of in-plane and out-of-plane fiber waviness in carbon fiber reinforced composites. Composite Structures 2022; 297:115932.

[28] Hsiao HM, Daniel IM. Effect of fiber waviness on stiffness and strength reduction of unidirectional composites under compressive loading. Composites Science and Technology. 1996; 56:581–593.

[29] Takeda T. Micromechanics model for three-dimensional effective elastic properties of composite laminates with ply wrinkles. Composite Structures. 2018; 189:419–427.

[30] Wen A, Ma L, Shen C, Guo J, Zheng J. Wrinkle defects investigation on displacement response in fiber-reinforced composites using grating projection measurement. Composite Structures 2022; 297:115867.

[31] El-Hajjar RF, Petersen DR. Gaussian function characterization of unnotched tension behavior in a carbon/epoxy composite containing localized fiber waviness. Composite Structures. 2011; 93:2400–2408.